\pdfoutput=1

\documentclass[11pt]{article}

\usepackage{EMNLP2022}

\usepackage{times}
\usepackage{latexsym}

\usepackage[T1]{fontenc}

\usepackage[utf8]{inputenc}

\usepackage{microtype}

\usepackage{inconsolata}
\usepackage{amsfonts}
\usepackage{amsmath}
\usepackage{booktabs}
\usepackage{graphicx}
\usepackage{comment}
\input{Definitions}

\usepackage{todonotes}

\newcommand{\ourmodel}{\texttt{CORE}}

\newcommand{\bert}{BERT}

\newcommand{\clstoken}{\texttt{[CLS]}}

\newcommand\blfootnote[1]{%
  \begingroup
  \renewcommand\thefootnote{}\footnote{#1}%
  \addtocounter{footnote}{-1}%
  \endgroup
}

\title{Open-domain Question Answering via Chain of Reasoning over Heterogeneous Knowledge}

\author{
Kaixin Ma\textsuperscript{$\clubsuit$}$\dagger^*$,
Hao Cheng\textsuperscript{$\spadesuit$}$^*$,
Xiaodong Liu\textsuperscript{$\spadesuit$},
Eric Nyberg\textsuperscript{$\clubsuit$},
Jianfeng Gao\textsuperscript{$\spadesuit$}
 \\ 
  \textsuperscript{$\clubsuit$} Carnegie Mellon University
  \textsuperscript{$\spadesuit$} Microsoft Research 
 \\
  {\tt \{kaixinm,ehn\}@cs.cmu.edu}
  {\tt \{chehao,xiaodl,jfgao\}@microsoft.com}
}

\begin{document}
\maketitle
\begin{abstract}

We propose a novel open-domain question answering (ODQA) framework for answering single/multi-hop questions across heterogeneous knowledge sources.
The key novelty of our method is the introduction of the intermediary modules into the current retriever-reader pipeline.
Unlike previous methods that solely rely on the retriever for gathering all evidence in isolation,
our intermediary performs a chain of reasoning over the retrieved set.
Specifically, our method links the retrieved evidence with its related global context into graphs and organizes them into a candidate list of evidence chains.
Built upon pretrained language models, our system achieves competitive performance on two ODQA datasets, OTT-QA and NQ, against tables and passages from Wikipedia.
In particular, our model substantially outperforms the previous state-of-the-art on OTT-QA with an exact match score of 47.3 (45 $\%$ relative gain). \blfootnote{$\dagger$ Part of this work is done during an internship at Microsoft Research} \blfootnote{$*$ Equal contribution}
\end{abstract}

\section{Introduction}

\begin{figure*}
    \centering
    \includegraphics[scale=0.48]{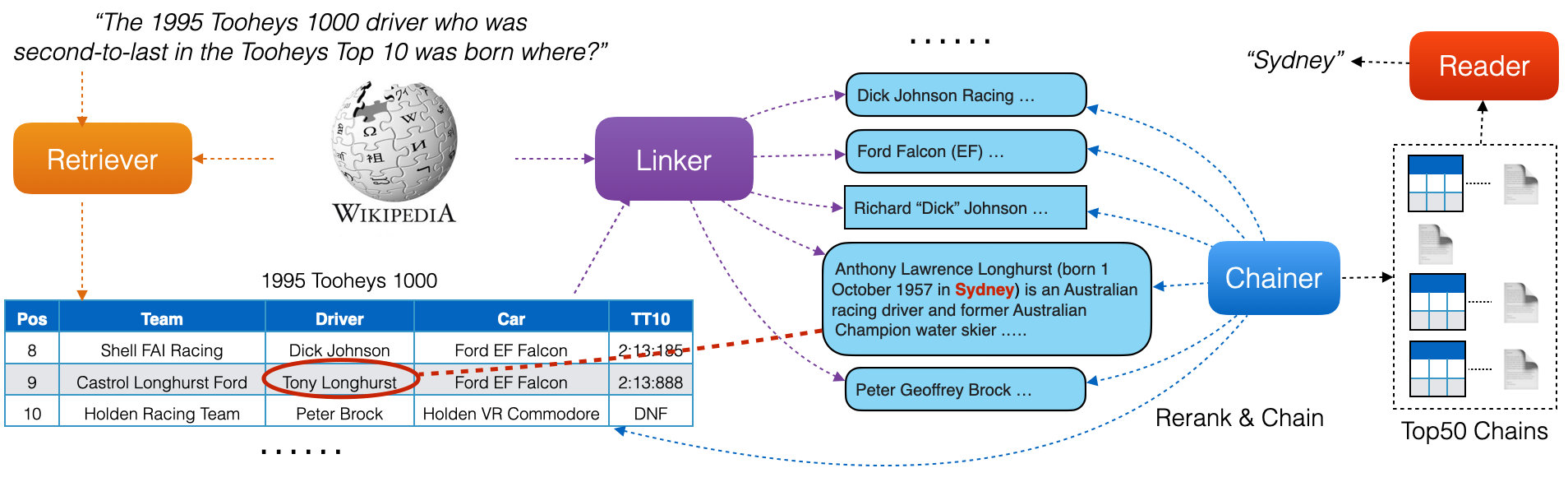}
    \caption{An illustration of \ourmodel\ for ODQA. 
    Given a question, the \textit{retriever} first finds hop-1 evidence from the entire Wikipedia (orange arrows). 
    Then the \textit{linker} gathers relevant documents for the hop-1 evidence (purple arrows), which are treated as hop-2 evidence.
    Next, the \textit{chainer} reranks all hop-1 and hop-2 evidence and splices them together into evidence chains (blue arrows).
    Finally, the \textit{reader} takes in the top-50 chains and produces the answer (black arrows).
    The gold evidence chain is marked in red.}
    \label{fig:model}
\end{figure*}

The task of open-domain question answering (ODQA) typically involves multi-hop reasoning, such as finding relevant evidence from knowledge sources, piecing related evidence with context together, and then producing answers based on the final supportive set.
While many questions can be answered by a single piece of evidence \cite{joshi-etal-2017-triviaqa,kwiatkowski-etal-2019-natural},
answering more complex questions are of great interest and require reasoning beyond \textit{local} document context \cite{yang-etal-2018-hotpotqa, geva-etal-2021-aristotle}.
The problem becomes more challenging when evidence is scattered across heterogeneous sources, \eg unstructured text and structured tables \cite{chen2020open}, which necessitates hopping from one knowledge modality to another. 
Consider the question in \autoref{fig:model}.
To form the final answer, a model needs to find the entry list (a table) of the mentioned touring car race, look up the driver name with the correct rank, search for the corresponding driver information, and extract the birthplace from the free-form text.

\begin{figure}
    \centering
    \includegraphics[scale=0.5]{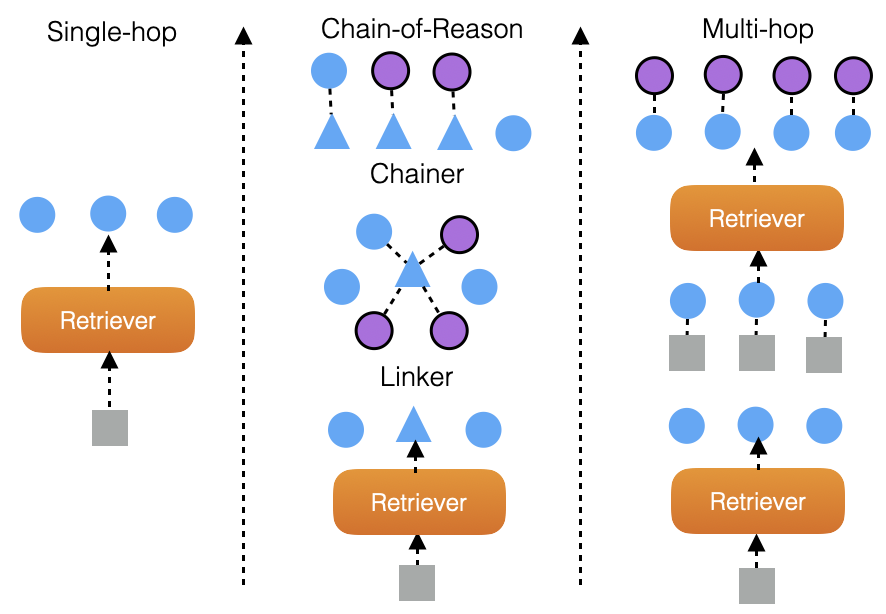}
    \caption{Our \ourmodel\ vs. previous retriever-only methods for evidence discovery. The grey square is the question, the blue dots are 1st-hop passages, the blue triangles are 1st-hop tables, and the purple dots are 2nd-hop passages.}
    \label{fig:concept}
\end{figure}

Existing \textit{retriever-reader} methods \citep[][\textit{inter alia}]{min2021neurips} for ODQA mainly customize the retriever model for tackling individual question types, \ie exclusively relying on the retriever to gather all necessary context in a query-dependent fashion.
As shown in \autoref{fig:concept}, the single-hop model \cite{karpukhin-etal-2020-dense} only retrieves a list of isolated passages (blue dots).
For multi-hop cases, an iterative retriever looks for a query-dependent path of passages (blue-purple dot chains) \cite{xiong2020answering}, \ie the later hop passages are retrieved using expanded queries including the original question and previous hop passages.
Although those retrieval-only methods achieve promising results on their targeted cases, the customized retrievers are unable to generalize well.
For example, an iterative passage retriever trained with both multi-hop and single-hop questions performs poorly over both types \cite{xiong2020answering}.
For real-world applications with heterogeneous knowledge sources, it is desirable for an ODQA system to handle both cases well, and the retrieval-only methods are unlikely to succeed.

We propose a novel \textbf{C}hain \textbf{O}f \textbf{RE}asoning (\ourmodel) framework that generalizes well on both single-hop and multi-hop question answering.
The main contribution of \ourmodel\ is the introduction of two intermediary modules, the linker and the chainer, that play the bridging role between the retriever and the reader, \ie 
piecing together related evidence with necessary context for single/multi-hop questions.
These two modules work in a forward-backward fashion.  
In the forward pass, the linker, a novel table entity linking model (\S\ref{subsec:linker}), links the raw evidence with its related context (\eg table-passage graphs in \autoref{fig:concept}). 
The chainer, a new unsupervised reranker (\S\ref{subsec:chainer}), then prunes all linked evidence using the corresponding question generation scores from a pretrained language model \cite{sanh2021multitask} to form a shortlist of relevant evidence chains in a backward noisy channel fashion (\eg table-passage paths in \autoref{fig:concept}).
By delegating the \textit{hopping} operation to the intermediary, our formalization can potentially gather evidence more effectively over different question types.


To demonstrate the effectiveness of \ourmodel, we evaluated the proposed model on two popular ODQA datasets, the multi-hop dataset OTT-QA \cite{chen2020open} and the single-hop dataset NQ \cite{kwiatkowski-etal-2019-natural}.
Empirically, we show that our approach is general for both types of reasoning in ODQA.
In particular, \ourmodel\ substantially outperforms the previous state-of-the-art (SOTA) on OTT-QA by 14+ points on exact match scores ($45\%+$ relative gain), and it is competitive with SOTA models on NQ. Moreover, we show that one single unified model can learn to solve both tasks under our framework. 
From our analysis, we also find that our evidence chains can potentially help answer single-hop questions by enriching the evidence with more supportive context. \footnote{Data and code available at \url{https://github.com/Mayer123/UDT-QA}}

\section{Overview of the \ourmodel\ Framework}

The \ourmodel\ framework is designed to answer questions 
where the answer is a contiguous span from a table $t$ or a passage $p$.
Here neither $t$ nor $p$ is given, so they need to be retrieved from the table corpus $\mathbb{C}_{t}$ and the passage corpus $\mathbb{C}_{p}$, respectively.
For single-hop questions, a single $t$ or $p$ may be sufficient,
whereas for multi-hop questions, one or more $t$ and $p$ are required to find the answer. 

As shown in \autoref{fig:model}, \ourmodel\ consists of a \textit{retriever}, 
a \textit{linker}, a \textit{chainer} and a \textit{reader}. 
We adopt the DPR model \cite{karpukhin-etal-2020-dense} as our retriever.
We only briefly describe the retriever here as it is not the main focus of our work.
The DPR is a bi-encoder model that consists of a question encoder and a context encoder.
For our setting, the questions and passages/tables are represented by the \clstoken~embedding produced by their respective encoder, and the retrieval is done by maximum inner product search in the vector space.
For a given question, we use DPR to retrieve the initial evidence set which includes tables and passages. 

Given the initial evidence set (\eg the car race entry list table in \autoref{fig:model}), our intermediary module produces a list of query-dependent evidence chains (\eg the red line linked evidence chain consisting of the car race entry list and the driver's Wikipedia page).
We first propose a linker model (\S\ref{subsec:linker}) to expand the candidate evidence set by including extra passages related to tables in the initial set (purple arrows in \autoref{fig:model}).
This step allows the model to enrich the evidence context, especially including reasoning chains needed for multi-hop questions. 
Since there could be many links between a piece of evidence and others (\ie a densely connected graph), considering all links is computationally infeasible for the downstream reader.
Thus, we develop a chainer model (\S\ref{subsec:chainer}) to prune the evidence graph with the corresponding question and then chain the evidence across hops together to form query-dependent paths.
Here, we only keep top-$K$ scored chains for reading so that the reader can work on a fixed computation budget.

Finally, the Fusion-in-Decoder (FiD) \cite{izacard-grave-2021-leveraging}, a T5-based generative model \cite{raffel2019exploring}, is used as the reader for generating the final answer.
The model first encodes each top-$K$ evidence chain independently along with the question.
During decoding, the decoder can attend to all chains, thus fusing all the input information. 

\section{Intermediary Modules}
In this part, we present the two key components of \ourmodel\ for supporting multi-hop reasoning, \ie the linker for building evidence graphs and the chainer for forming query-dependent paths.

\subsection{Linker}
\label{subsec:linker}
In this work, we mainly focus on linking an entity mention in the retrieved evidence to the corresponding Wikipedia page for building evidence graphs.
This setup is related to the recent entity linking work \cite{wu-etal-2020-scalable}. However, there are important modifications for ODQA.
In particular, instead of assuming the entity mention as a prior,
we consider a more realistic end-to-end scenario for ODQA: the linker model has to first \textit{propose candidate entity mentions (spans)} for a given evidence
(\eg ``Tony Longhurst'' in \autoref{fig:model}),
and then \textit{links the proposed mention} to its Wikipedia page.
Another major difference is that we study entity mentions in tables instead of text.
As tables contain more high-level summary information than text, using tables as pivots for constructing evidence graphs can potentially help improve the recall of evidence chains for QA. In the meanwhile, this task is challenging due to the mismatch between the lexical form of the table cells and their linked passage titles. For example, the table of "NCAA Division I women's volleyball tournament" contains the cell \textit{VCU}, which refers to \textit{VCU Rams} instead of \textit{Virginia Commonwealth University}. Thus simple lexical matching would not work. 

In the following, we first describe the model for entity mention proposal and then present a novel entity linking model for mentions in tables.
Both models are based on a pretrained language model, BERT \cite{devlin-etal-2019-bert}.
Following previous work \cite{oguz2020unikqa}, we flatten the table row-wise into a sequence of tokens for deriving table representations from BERT.
In particular, we use $x_1, \ldots, x_N$ to denote an input sequence of length $N$.
Typically, when using BERT, there is a prepended token \clstoken~for all input sequences, \ie $\clstoken, x_1, \ldots, x_N$.
Then the output is a sequence of hidden states $\hvec_\clstoken, \hvec_1, \ldots, \hvec_N\in\RR^d$ from the last BERT layer for each input token, where $d$ is the hidden dimension.

\noindent \textbf{Entity Mention Proposal} In realistic settings, the ground truth entity mention locations are not provided.
Directly applying an off-the-shelf named entity recognition (NER) model can be sub-optimal as the tables are structured very differently from the text.
Thus, we develop a span proposal model to label the entity mentions in the table.
Specifically, we use BERT as the encoder (\bert\textsuperscript{m}) and add a linear projection to predict whether a token is part of an entity mention for all tokens in the table,
\begin{align}
    \hvec\textsuperscript{m}_1, \ldots, \hvec\textsuperscript{m}_N & = \bert\textsuperscript{m}(t_1, \ldots, t_N), \\
    \hat{\yvec} & = W \hvec\textsuperscript{m},
\end{align}
where $\hvec\textsuperscript{m} \in \RR^{N \times d}$ and $W\in \mathbb{R}^{2 \times d}$.
The model is trained with a token-level binary loss 
\begin{equation}
    -\frac{1}{N}\sum^{n=1}_N
   (y_n \log P(\hat{\yvec})_1  + (1 - y_n)\log P(\hat{\yvec})_0),
\end{equation}
where $y_n$ is the 0-1 label for the token at position $n$, and $P(\cdot)$ is the softmax function.

\noindent\textbf{Table Entity Linking}
Once the candidate entity mentions are proposed, we follow \citet{wu-etal-2020-scalable} to use a bi-encoder model for linking.
Similarly, two BERT models are used to encode tables (\bert\textsuperscript{t}) and passages (\bert\textsuperscript{p}), respectively.
In contrast, as there are multiple entity mentions for each table, we want to avoid repetitively inserting additional marker tokens and re-computing representations for each mention occurrence.
Accordingly, we cannot simply take the \clstoken~ embeddings for linking as previous work \cite{wu-etal-2020-scalable}.
Inspired by \citet{baldini-soares-etal-2019-matching}, to represent each entity mention, we propose a new entity representation based on the entity start and end tokens.
For an entity mention with a start position $i$ and end position $j$ in the table, we compute our proposed entity embedding $\evec\in\RR^d$ for linking by
\begin{align}
    \evec = & (\hvec\textsuperscript{t}_i + \hvec\textsuperscript{t}_j)/2, 
\end{align}
For passages, we directly take the \clstoken~ hidden state $\pvec=\hvec\textsuperscript{p}_\clstoken\in\RR^d$ as the passage representation.
Following the literature, the table entity linker is trained based on a contrastive learning objective
\begin{equation}
      L_\textsubscript{sim} = -{\exp(\text{sim}(\evec, \pvec^+)) \over \sum_{\pvec^\prime\in\mathcal{P}\cup\{\pvec^+\}} \exp(\text{sim}(\evec, \pvec^\prime))},
    \label{eqn:obj_sim}  
\end{equation}
where $\pvec^+$ is the corresponding (positive) passage and $\mathcal{P}$ is the irrelevant set of negative passages.

\noindent \textbf{Training}
To train our linker including entity mention proposal and table entity linking, we leverage a small set of Wikipedia tables with ground truth hyperlinks, where each linked mention and the first paragraph of the linked page constitute a positive pair (see \autoref{sec:appendix-training}).
Similar to \citet{wu-etal-2020-scalable}, we use BM25 \cite{bm25} to mine hard negative pairs for each entity mention.

\noindent \textbf{Inference} During inference, we first use the entity span proposal model to label entity mentions in the tables and then run the table entity linking model to link predicted mentions to passages via maximum inner product search \cite{johnson2019billion}. Here, we allow searching over all passages rather than the first paragraph of each Wikipedia page for recall purposes.
For efficient inference, the passage corpus can be pre-computed and stored as done in dense retrieval \cite{karpukhin-etal-2020-dense}. Then we can build links for all tables in the corpus (inducing graph-structured corpus), and directly use the graph to enrich the isolated evidence items from the retriever. 

\subsection{Chainer}
\label{subsec:chainer}
Although the linker model can effectively enrich the table with all relevant passages and provide possible hopping paths for reasoning,
the amount of resulting information is too overwhelming if sent directly for reading, \ie a table could contain multiple entity mentions, resulting in a densely connected evidence graph.
To make it easy for the reader, we use the chainer model to prune graphs and extract the most relevant paths.
Since the linker (\S\ref{subsec:linker}) builds the evidence graphs in a query-agnostic fashion, an important function of chainer here is to incorporate the question when selecting the top-K evidence paths for answer inference.

Motivated by recent work \cite{sachan2022improving} in using pretrained generative language models for passage reranking, we also build the chainer on T0 \cite{sanh2021multitask} in a zero-shot fashion, \ie no training is required.
Different from their approach, we design a novel relevance scoring for query-dependent hybrid evidence path reranking rather than isolated passages.

Given a question $q$, we model the relevance of a question-table-passage path using the following conditional
\begin{align}
    P(t, p|q) = P(p|t, q)P(t|q),
    \label{eqn:relevance_orig}
\end{align}
where $t \in \mathbb{C_T}$, $p \in \mathbb{C_P}$.
The second term $P(t|q)$ is modeled by our retriever. Given that our linker is query-agnostic (\ie only modeling $P(p|t)$), we do not have a good estimation for $P(p|t,q)$ on the right-hand side.
To remedy this, we apply the Bayes rule to \autoref{eqn:relevance_orig}
\begin{equation}
    \label{eqn:relevance_score}
    P(t, p|q) \approx P(q|t, p)P(p|t)P(t|q).
\end{equation}
To estimate $P(q|t, p)$, we use the question generation likelihood as in \citet{sachan2022improving}. 

Different from \citet{sachan2022improving}, here we have two conditional variables. Naively computing question generation scores on all pairs results in quadratic complexity which is very computation intensive for T0.\footnote{Even the smallest T0 model has 3B parameters.} 
To reduce the inference cost, we further decompose $P(q|t,p)$ into two question generation scores $S_{T0}(q|p)$ and $S_{T0}(q|t)$, both based on the question generation likelihood from T0.
In this way, we can reuse $S_{T0}(q|t)$ for corresponding linked passages with a linear cost.
To compute $S_{T0}(q|p)$ and $S_{T0}(q|t)$, we append the instruction \textit{``Please write a question based on this passage.''}\footnote{Changing ``passage'' to ``table'' in the prompt does not make much difference.}
to every passage/table and use the same mean log-likelihood of the question tokens conditioned on the passage/table  \cite{sachan2022improving}. 

As our pilot study suggests that the query-agnostic linker scores are not so informative for query-table-passage paths,
we only combine the retriever score with two question generation scores from \autoref{eqn:relevance_score} as the final chainer score for reranking evidence paths (\ie $P(p|t)$ is dropped)
\begin{equation}
\label{eqn:chainer}
 S_{R}(t, q) + \alpha S_{T0}(q|t) + \beta S_{T0}(q|p),
\end{equation}
where $S_{R}(t, q) \sim P(t|q)$, and is defined as
\begin{equation}
    S_{R}(t, q) = -\text{log} (\frac{\exp(\text{sim}(t, q))}{\sum_{t_i\in\mathcal{T}} \exp(\text{sim}(t_i, q))}),
\end{equation}
$\text{sim}(\cdot, \cdot)$ is the unnormalized retriever score, $\mathcal{T}$ is the first-hop evidence set,
$\alpha$ and $\beta$ are hyper-parameters.
For singleton cases (first-hop table/passage without link), we modify the last two terms of \autoref{eqn:chainer} to $2\alpha S_{T0}(q|t)$ and $2\alpha  S_{T0}(q|p)$ for tables and passages, respectively.
This can help ensure that the chainer scores for singletons and table-passage paths are on the same scale.  
Then we sort all singletons and chains using the chainer score and keep the top-k results.
We also use heuristics to reduce potential duplication and details are in \autoref{sec:appendix-chain}. 

\section{Experiments}
In this section, we first describe datasets and knowledge sources (\S\ref{ssec:datasets}).
Then we discuss single-set and joint-set experiment setups (\S\ref{ssec:exp_setup}) and baselines for comparison (\S\ref{ssec:baselines}).
Finally, we present experiment results on OTT-QA and NQ (\S\ref{ssec:main_results}).

\subsection{Datasets}
\label{ssec:datasets}
\textbf{OTT-QA} \cite{chen2020open} is an open-domain QA dataset that contains mostly multi-hop questions. 
These questions require joint reasoning over evidence from tables and text passages. \\
\noindent \textbf{Natural Questions} (NQ) \cite{kwiatkowski-etal-2019-natural} contains real user queries submitted to the Google search engine and the questions are mostly single-hop and considered solvable using either a single text passage or a table. 
We adopt the open-domain setting proposed by \citet{lee-etal-2019-latent}. 

\noindent For OTT-QA, we adopt its official release of text passages and tables.
The passages are first paragraphs from all Wikipedia pages, whereas tables are well-formed tables from full Wikipedia (\ie no infoboxes, no multi-column/multi-row tables, etc.).
For NQ, we adopt the Wikipedia passage splits from \citet{karpukhin-etal-2020-dense} and we use the processed table sets of full Wikipedia released by \citet{ma-etal-2022-open}.
Statistics can be found in \autoref{sec:appendix-data}.

\subsection{Experimental Settings}
\label{ssec:exp_setup}
We train a single \textbf{linker} on the OTT-QA training set, and directly apply it for both tasks. For \textbf{chainer}, we apply the off-the-shelf T0-3B \cite{sanh2021multitask} for reranking on both tasks, \ie no updates to the model. Hence both models are task-independent. 

For both \textbf{retriever} and \textbf{reader}, we consider both \textit{single-set} setup where separate models are trained for individual tasks and the \textit{joint-set} setting where a single model is trained to solve both tasks. 
During inference, the retriever searches over task-specific knowledge sources.
For OTT-QA, since most of its questions have tables as the first-hop evidence, we run the retriever only on the table set in the first hop to find top-100 tables.
For NQ, the retriever searches over the joint index of all text and tables and keeps the top-100 items in the first hop. 
After retrieval, we use the linker to expand tables in the top-100 set into graphs and use the chainer to select the top-50 chains.  
As the chained evidence is typically longer, unless otherwise specified, we only use the top-50 chains 
and set the maximum sequence length to 500 for the FiD reader. 
More training details are in \autoref{sec:appendix-training}.

\begin{table}
\centering
\resizebox{\linewidth}{!}{
\begin{tabular}{lcccc}
\toprule
 & \multicolumn{2}{c}{Dev} & \multicolumn{2}{c}{Test} \\
 & \textbf{EM} &  \textbf{F1} & \textbf{EM} &  \textbf{F1}\\
\midrule
HYBRIDER \cite{chen-etal-2020-hybridqa} & 10.3 & 13.0 & 9.7 & 12.8 \\
FR+CBR\cite{chen2020open}  & 28.1 & 32.5 & 27.2 & 31.5 \\
CARP \cite{zhong2022reasoning} & 33.2 & 38.6 & 32.5 & 38.5 \\
\midrule
Oracle Link + FR+CBR & 35.2 & 39.9 & 35.0 & 39.5 \\
Oracle Link\&table + HYBRIDER & 44.1 & 50.8 & 43.0 & 49.8 \\
\midrule
\ourmodel\, (single) & \bf 49.0 & \bf 55.7 & 46.7 & 53.5 \\
\ourmodel\, (joint) & \bf 49.0 & \bf 55.7 & \bf 47.3 & \bf 54.1 \\
\bottomrule
\end{tabular}
}
\caption{End-to-end QA results on OTT-QA.}
\label{tab:ott_res}
\end{table}

\subsection{Baselines}
\label{ssec:baselines}
We briefly describe the baselines for both tasks.
For OTT-QA, the HYBRIDER \cite{chen-etal-2020-hybridqa} is a reading comprehension model for joint reasoning over tables and passages.
This baseline uses BM25 to retrieve relevant tables and passages.
Instead, Fusion Retriever + Cross-Block Reader (FR+CBR) \cite{chen2020open} first links table rows to passages using BM25 to build an index of linked documents (blocks). Then it trains a biencoder dense retriever (further enhanced by Inverse Cloze Task pretraining \cite{lee-etal-2019-latent}) to find the most relevant blocks. 
Then they use the ETC \cite{ainslie-etal-2020-etc} as the reader to process up to 4K tokens returned by the retriever.
CARP \cite{zhong2022reasoning} employs similar retriever and reader models as FR+CBR, and it additionally extracts hybrid knowledge chains to facilitate the reader's reasoning process. 
All approaches are dataset-specific, hence are unlikely to handle other question types from NQ. We note that both FR+CBR and CARP's reader components adopt models specifically designed for handling long sequences. Since our overall goal is to build a unified system for both single-hop and multi-hop questions, we choose FiD based on its ability to handle different cases. 
We also compare against two oracle settings from \citet{chen2020open}.
The first one uses gold hyperlinks in fusion retriever instead of BM25 to link table rows and passages. 
The second setting adopts gold hyperlinks and gold tables for the HYBRIDER reader, \ie no retrieval is involved.
This setting is previously considered as an estimated upper bound for this task \cite{chen2020open}. 

On NQ, we compare with text-only baselines: DPR \cite{karpukhin-etal-2020-dense} which applies a BERT-based reader to first select the best passage from top-k returned by the retriever and then extracts the answer span in it; FiD \cite{izacard-grave-2021-leveraging}; and UnitedQA \cite{cheng-etal-2021-unitedqa}) which is an extractive model based on ELECTRA \cite{clark2020electra} and enhanced with additional training objectives \cite{cheng-etal-2021-posterior,cheng-etal-2020-probabilistic}.
We also compare with models that consider tables as knowledge sources: Unik-QA \cite{oguz2020unikqa} which augments the document index with NQ tables and uses FiD reader to generate the answer; UDT-QA \cite{ma-etal-2022-open} which incorporates the tables from full Wikipedia and adopts UnitedQA as its reader. 

\begin{table}
\centering
\resizebox{\linewidth}{!}{
\begin{tabular}{lcc}
\toprule
 & \textbf{Tables} & \textbf{EM} \\
\midrule
DPR \cite{karpukhin-etal-2020-dense} & N & 41.5 \\
FiD \cite{izacard-grave-2021-leveraging} & N & 51.4 \\
UnitedQA \cite{cheng-etal-2021-unitedqa} & N & 51.8 \\
\midrule
Unik-QA \cite{oguz2020unikqa} & Y & 54.1 \\
UDT-QA \cite{ma-etal-2022-open} & Y & \bf 54.7 \\
\midrule
\ourmodel\, (single) & Y & 54.6 \\
\ourmodel\, (joint) & Y & 53.9\\
\bottomrule
\end{tabular}
}
\caption{End-to-end QA results on NQ test.}
\label{tab:nq_res}
\end{table}

\subsection{Results}
\label{ssec:main_results}
The end-to-end results on OTT-QA are shown in \autoref{tab:ott_res}. 
Our \ourmodel\,models substantially outperform all baselines by a large margin, illustrating the effectiveness of our proposed framework.
It is worth noting that our model also outperforms two oracle settings proposed in \citet{chen2020open}.
This is potentially because the capacity of their reader models is quite limited compared to ours.
For FR+CBR, the model can only read up to 4K tokens, whereas in our case, the FiD reader can process up to 25K tokens.
We also observe that evidence to questions in OTT-QA can be found in many different reasoning chains.
In other words, tables that are not annotated as gold may still provide valuable information for reasoning.
However, only one table is considered in their oracle experiment with HYBRIDER.
Also, it is worth noting that the joint model outperforms the single model on OTT-QA, indicating that our framework can effectively leverage NQ data to benefit OTT-QA.

\autoref{tab:nq_res} summarizes the results on NQ.
Similar to previous work using tables as the extra knowledge source, we also find our method to be consistently better than text-only baselines. 
Overall, \ourmodel\ achieves competitive performance compared to SOTA models. 
It is also worth noting that both Unik-QA and UDT-QA use the iterative training strategy for the retriever, which leads to higher retriever performance. In particular, UDT-QA achieves 91.9 recall@100 on the NQ test set, whereas we only get 90.3. This difference in retriever recall likely explains the gap in the end-to-end QA performance.  
Since iterative training of the retriever is not used by baselines on OTT-QA, we leave that out in our experiments. For consistency, we only train our retriever for one round. Also, we note that the joint model performs slightly worse than the single-dataset model, which is different from the trend observed on OTT-QA. We hypothesize that this is due to the distribution difference of the evidence chains in the two datasets. Most of the top-ranked evidence are chained cases on OTT-QA, whereas that reduces to only 30\% on NQ (the rest 70\% are singleton cases). The current FiD reader may have a hard time reasoning over both singleton and chained cases simultaneously. We leave further exploration for future work. 
\section{Analysis}
In this section, we conduct ablation studies and analyze different components of our framework.

\subsection{Ablation Study}
\begin{table}[!t]
\centering
\resizebox{\linewidth}{!}{
\begin{tabular}{lccc}
\toprule
 & \textbf{\#Docs} &  \textbf{Max Length} & \textbf{EM} \\
\midrule
\ourmodel\, (single) & 50 & 500 & 49.0  \\
no QGS of hop1 & 50 & 500 & 45.1  \\
no QGS of hop1 & 100 & 300 & 44.8\\
no QGS of hop1\&2 & 100 & 300 & 38.7  \\
no Chainer & 100 & 300 & 29.1 \\
\bottomrule
\end{tabular}
}
\caption{Chainer ablation on OTT-QA dev.}
\label{tab:ott_ablation}
\end{table}
We start by ablating our chainer on the OTT-QA task using \ourmodel\,(single) from \autoref{tab:ott_res}.
The results are summarized in \autoref{tab:ott_ablation}.
First, we experiment with
\textbf{removing question generation score (QGS) for hop1 evidence} (row 2).
In this case, \autoref{eqn:chainer} will not have the second term and a large performance drop is observed, suggesting that it is important to apply chainer to reweight the retrieved items.
Under the same setting, we try to increase the number of items and decrease the max sequence length for reading (row 3).
Here we are interested in seeing the impact of reading more chains while constraining the length of individual chains.
The slight performance drop indicates that potential information loss from longer chains is detrimental to the final performance. 

Then we test \textbf{removing QGS entirely} (row 4).
As there is no reranking at all, we simply take all linked passages for each table based on the initial retrieved order.
Here, since we exhaustively include all linked passages for each table, the reader budget can be quickly filled by the top few retrieved tables, which probably makes the reader suffer from information loss with too many irrelevant items.
Indeed we see another large drop in the QA performance, showing the importance of reranking.
Finally, we experiment with \textbf{removing chainer} (row 5).
It goes one step further by not concatenating the table rows with linked passages and only passing each piece of evidence independently.
Here we are interested in whether FiD is able to fuse the information without explicit chaining.
The largest performance drop is observed here, suggesting that chaining the items from different hops is vital for the reader. 

We also study the effect of the number of retrieved items as input to the reader on its QA performance by varying \textbf{the number of chains sent to the reader}.
The results on OTT-QA dev set are shown in \autoref{tab:ott_res_length}.
As we can see, when reading only the top 10 chains (much less than 4K tokens on average considered by previous work \cite{chen2020open,zhong2022reasoning}), \ourmodel\ still outperforms the previous SOTA method by a large margin, further validating the effectiveness of our framework.
In summary, the reader performance increases with more items, and we hypothesize that the reader can be further improved when given a larger capacity.
We leave this exploration for future work.

\begin{table}[t!]
\centering
\small
\begin{tabular}{lccc}
\toprule
 \textbf{\# Docs} & \textbf{Avg \# Tokens} & \textbf{EM} &  \textbf{F1}\\
\midrule
 50 & 11,897 & \bf 49.0 & \bf 55.7 \\
 20 & 4,757 & 44.9 & 51.8 \\
 10 & 2,351 & 40.8 & 47.0 \\
\bottomrule
\end{tabular}
\caption{Reader ablation with different number of documents on OTT-QA dev.}
\label{tab:ott_res_length}
\end{table}

\subsection{Impact of Linker \& Chainer}
We measure evidence recall scores of our joint model to better understand the impact of the linker and the chainer.
On OTT-QA, since most questions require multi-step reasoning, we evaluate the retriever by both gold cell recall and answer recall.
Here the gold cells refer to table cells whose gold linked passage contains the answer.
We consider a table chunk to be gold if it contains at least one gold cell.
For answer recall, we use top-K chains produced by the chainer.
Since OTT-QA official test set is hidden, we only report dev set results in \autoref{tab:ott_retri}.
As expected, the answer recall scores are extremely low without the linker and chainer,
as most OTT-QA questions are multi-hop and the answers are likely to appear in the hop-2 passages.
To verify if the retriever itself is able to discover the full reasoning chain, we further allow the retriever to search through the joint index of tables and passages (last row).
Compared with the first row, there is some improvement.
On the other hand, with the help of the linker and chainer (row 2), AR@20 and AR@50 
are substantially improved, which also explains our superior QA performance.
We also observe similar trends on NQ, and more results and discussion are in Appendix~\ref{subsec:appendix-linker-chainer}. 

\begin{table}[t!]
\centering
\resizebox{\linewidth}{!}{
\begin{tabular}{lcccc}
\toprule
 & \textbf{R@20} &  \textbf{R@100} & \textbf{AR@20} &  \textbf{AR@50}\\
\midrule
Joint Retriever & \bf 83.8 & \bf 92.1 & 31.8 & 37.6 \\
+Linker\&Chainer & 83.5 & 90.8 & \bf 74.5 & \bf 82.9 \\
\midrule
Retrieve full index & 82.4 & 90.6 & 34.6 & 42.6 \\
\bottomrule
\end{tabular}
}
\caption{Evidence recall of the joint retriever on OTT-QA dev set, where R@K evaluates gold table chunk recall and AR@K evaluates answer recall.}
\label{tab:ott_retri}
\end{table}

\begin{table}[t!]
\centering
\resizebox{\linewidth}{!}{
\begin{tabular}{lcccc}
\toprule
 & \textbf{\# Link} & \textbf{\# No Link} & \textbf{AR@20} &  \textbf{AR@50}\\
\midrule
 \ourmodel\, & 2,214 & 0 & 74.5 & 82.9 \\
 Classifier & 2,133 & 81 & 73.3 & 81.9 \\
 \midrule
 \ourmodel\, & 8,757 & 0 & 85.7 & 88.1 \\
 Classifier & 370 & 8,387 & 86.0 & 88.1 \\
\bottomrule
\end{tabular}
}
\caption{Answer recall on OTT-QA (top) and NQ (bottom) dev with different linking strategies.}
\label{tab:q_type}
\end{table}

\begin{table*}[h!]
\small
\centering
\resizebox{\linewidth}{!}{
\begin{tabular}{ll}
\toprule
 \textbf{Q\&A}& \textbf{Evidence Chain} \\
\hline
\textbf{Q:} who plays ryders mum & \textbf{Table Title:} List of Home and Away characters \\
on home and away & || character, actor(s), duration || \textit{ryder jackson}, lukas radovich, 2017- || \textit{Ryder Jackson} is  \\
\textbf{A:} Lara Cox & a fictional character from the Australian television soap opera "Home and Away" ... \\
 & \textbf{His mother is Quinn Jackson (Lara Cox)}, who is estranged from Alf. When ...\\
\midrule
\textbf{Q:} what was blur 's first & \textbf{Table Title:} List of UK top-ten singles in 1994 \\
number one single in the uk & || artist, weeks, singles || blur, 3, \textit{"girls \& boys"}, "parklife" || \textit{Girls \& Boys} (Blur song) is a 1994 \\
\textbf{A:} Country House & song by British rock band Blur. It was released ... "Girls \& Boys" was Blur's first top 5 hit and \\
& their most successful single until \textbf{"Country House" reached number 1} the following year ... \\
\midrule
\textbf{Q:} Who is the dad of the cyclist  & \textbf{Table Title:} Holland Hills Classic \\
that placed directly behind & || Year, First, Second, Third || 2011, Marianne Vos, Marieke van Wanroij, \textit{Jessie Daams} || \\
Marieke van Wanroij at the & \textit{Jessie Daams} Jessie Daams ( born 28 May 1990 ) is a Belgian racing cyclist . \\
2011 Holland Hills Classic ? & She competed in the 2013 UCI women 's road race in Florence .  \\
\textbf{A:} Hans Daams & \textbf{Her father is the Dutch cyclist Hans Daams} . \\
\midrule
\textbf{Q:} What other team did the& \textbf{Table Title:}  2012 Charlotte Eagles season \\
Cuban player on the 2012 Char- & || No, Position, Player, Nation || 19, Midfielder, \textit{Miguel Ferrer}, Cuba || \textit{Miguel Ferrer} \\
lotte Eagles team play for ?  & (footballer) Miguel Ferrer ( born March 28 , 1987 ) is a Cuban footballer who \textbf{played for} \\
\textbf{A:} Wichita Wings & \textbf{the Wichita Wings} in the Major Indoor Soccer League . \\
\bottomrule
\end{tabular}
}
\caption{Example evidence chains found by our \ourmodel, where || separates tables and passages. The answer evidence is \textbf{bold} and the linked entity mention in the table is \textit{italic}. The first two are from NQ and the latter are from OTT-QA.}
\label{tab:cases}
\end{table*}

\subsection{Alternative Linking Strategy}
Since most NQ questions only require single-hop reasoning,
one alternative linking strategy for NQ is to skip the linker for single-hop ones.
As the question type is typically unknown in real cases,
we experiment with training a question classifier to predict whether a question requires multi-step reasoning or not.
We directly use the encoded question representation produced by our joint retriever, and train a linear classifier for this task.
For simplicity, we consider all NQ questions to be single-hop and all OTT-QA questions to be multi-hop.
Based on this, the classifier can achieve 95.9\% accuracy on the combined dev set of NQ and OTT-QA.
We then proceed to compute answer recall for both NQ and OTT-QA using the classifier to decide 
the linking strategy for each question.
The results are shown in \autoref{tab:q_type}.
We observe the difference in answer recall between \ourmodel\ and that using the question classifier 
is quite small.
Thus, future work can also use this strategy for handling different types of questions.

\subsection{Linker Performance}
\label{subsec:appendix-linker}
The success of our framework depends on the linking quality, thus we also evaluated our linker model as a standalone module to better understand its performance. Here we use the 789 tables in the OTT-QA dev set with ground truth hyperlinks for evaluation. For metrics, we compute precision, recall and F1 score for finding all the links for each table and only consider the top-1 retrieved results in all settings. We compare the cell linker model proposed by \citet{chen2020open} as a baseline. In this model, they first train a GPT-2 \cite{radford2019language} model on OTT-QA training set to generate queries for every cell in the table (empty if the model decides to not link a certain cell), and then use BM25 to retrieve passages. In addition, we also consider a baseline that uses ground truth table cells (cells that have links) as queries and retrieve with BM25. The results are shown in \autoref{tab:ott_dev_link}. Our linker achieves the best F1 score compared to the two baselines, and the advantage on recall is especially prominent. The oracle+BM25 model has the best precision because the information for whether a cell requires linking is given as a prior. However, it cannot retrieve well when the passage title does not overlap significantly with the cell text, as discussed in (\S\ref{subsec:linker}). The GPT2 model can alleviate this issue to some extent by generating additional terms for matching, however, it still lags behind our proposed linker model. 

\begin{table}[hbt!]
\centering
\resizebox{\linewidth}{!}{
\begin{tabular}{lccc}
\toprule
 & \textbf{Precision} &  \textbf{Recall} &  \textbf{F1}\\
\midrule
Oracle cell + BM25 & \bf 61.7 & 51.0 & 55.9 \\
GPT2+BM25 & 59.9 & 58.3 & 59.1 \\
Our Linker & 60.3 & \bf 63.0  & \bf 61.6 \\
\bottomrule
\end{tabular}
}
\caption{Linker variants on OTT-QA Dev set tables. In total, there are 20,064 unique passages attached to these 789 tables}
\label{tab:ott_dev_link}
\end{table}

\subsection{Case Study}
We manually inspect evidence chains found by our model to better understand the benefits of our intermediary modules.
Examples where predicted chains contain supportive evidence are in \autoref{tab:cases}. 

Despite that NQ questions are usually short and single-hop, we posit that some questions can potentially benefit from the proposed chain of reasoning.
As illustrated by the first two examples, though the entity mentions from the tables are not directly relevant, the table-passage chains actually contain the supportive information.
Therefore, in contrast to considering the evidence in isolation, our way of constructing query-dependent table-passage chains is likely to improve the density of relevant information for single-hop questions.

For the OTT-QA chains, there is little overlap between the question and the target entity passage as expected.
Thus, it is relatively difficult for the retriever to succeed in gathering all relevant information alone.
On the contrary, our framework can effectively handle it with our proposed linking and chaining operations.   
\section{Related Work}
Multi-hop question answering (QA) has been studied extensively for both knowledge base (KB) setting \cite{yih-etal-2015-semantic,zhang2017variational,talmor-berant-2018-web} and open-domain setting \cite{yang-etal-2018-hotpotqa,feldman-el-yaniv-2019-multi,geva-etal-2021-aristotle}. For KB-based setting, the models are trained to parse questions into logical forms that can be executed against KB \cite{das-etal-2021-case,yu2022decaf} or directly select entities in KB as answers \cite{sun-etal-2019-pullnet}. On the other hand, the open-domain setting requires the model to retrieve multiple pieces of evidence from a textual corpus and then produce the answer \cite{nie-etal-2019-revealing,zhu-etal-2021-adaptive}. Our work falls into the latter category.  

One stream of open-domain work uses multiple rounds of retrieval, where the later rounds depend on the previous ones.
It is typically achieved by some form of query reformulation, \eg expanding the question with previous passages \cite{xiong2020answering,zhao-etal-2021-multi-step} or re-writing the query using relevant evidence keywords \cite{qi-etal-2019-answering,qi-etal-2021-answering}.
The other line of work directly leverages the gold graph structure to expand the initial set of passages for hopping \cite{ding-etal-2019-cognitive,asai2019learning,10.1145/3404835.3462942}, assuming the presence of oracle hyperlinks.  
Different from those customized text-only multi-hop methods, our approach constructs evidence graphs on-the-fly and we show that it can handle single/multi-hop questions with a unified model over heterogeneous knowledge sources. Moreover, we do not presume the existence of gold hyperlinks in the corpus, making our model more applicable in realistic settings. 

There are also recent efforts in leveraging structured knowledge for ODQA.
\citet{li-etal-2021-dual} proposed to leverage information from both text and tables to generate answers and SQL queries.
\citet{oguz2020unikqa} studied the benefits of both tables and knowledge bases on a set of ODQA tasks.
\citet{ma-etal-2022-open} introduced a unified knowledge interface that first verbalizes tables and KB sub-graphs into text and then uses a single retriever-reader model to handle all knowledge sources. 
Similarly, we also consider heterogeneous knowledge sources for ODQA.
Instead of developing task-specific models and considering the evidence in isolation, we focus on finding evidence paths across different knowledge types for single/multi-hop questions.

\section{Conclusion}
In this paper, we present a new framework, \ourmodel, for ODQA over heterogeneous knowledge sources.
With the novel task-agnostic intermediary module, \ourmodel~can effectively handle single/multi-hop tasks using a unified model and achieve new SOTA results on OTT-QA.
Our analyses show that the intermediary module is necessary to achieve good results. 
For future work, it would be interesting to apply our framework to other complex reasoning tasks such as fact verification \cite{thorne-etal-2018-fever,aly2021feverous} or commonsense QA \cite{talmor2022commonsenseqa}. 
\clearpage

\section{Limitations}
We identify the limitations of our study below.

Conceptually, our proposed linker model is generic and it should be able to build edges between documents regardless of their type, e.g. passage-to-passage, passage-to-table. However, in this paper, we focus on one instantiation of the linker model, linking tables to passages. It would be interesting to build a more generic linker that is able to handle different edge types. 

Although we have experimented with OTT-QA and NQ, which exhibit very different question styles and reasoning types, both tasks are built on Wikipedia.
As most pretrained language models used in our work are also trained using Wikipedia, there are potential issues in generalization.
It would be interesting to apply our framework to other domains and test its generalizability, \eg biomedical domain. 

One major bottleneck we found in our experiments is computation. Since we use the T0-3B model (which is the cheaper one, T0-11B is the recommended model by \citet{sanh2021multitask}) as our chainer, it incurs a very large memory footprint and computation cost even just for inference.
Moreover, the FiD reader model is built on T5-large. Encoding top-50 chains is still computationally expensive.
In our case, even with a per GPU batch size of 1, the model still cannot run on V100-32G GPUs, thus we had to resort to gradient checkpointing, leading to longer running time.
It would be interesting to experiment with other alternatives that require less computation. 


\bibliography{anthology,custom}
\bibliographystyle{acl_natbib}

\clearpage
\appendix
\section{Chaining Strategy}
\label{sec:appendix-chain}
At the chaining step, a table $t$ can appear multiple times in this evidence chain list, if there are multiple passages linked to it. Also, a single passage can be linked to multiple tables, resulting in potential redundancy. Moreover, concatenating the table with the passage may make the sequence too long for the reader to handle. To reduce the duplication and sequence length, we adopt the following strategy: for a document in the first hop, we only add it to the Top-K list if it has not been included before; similarly, for a linked passage in the second hop, we only added it to the Top-K if it has not been added before, and we only concatenate the table header and its entity mention's row to it, and then include it as a separate document. We iterate over the reranked list of chains until reaching K chains.  

\section{Dataset Statistics}
\label{sec:appendix-data}
The dataset statistics are shown in \autoref{tab:stats}. For both datasets, we follow \citet{ma-etal-2022-open} to split tables into chunks of approximately 100 words, which results in about two chunks for each table. Thus the size of the table index also increased. (840K table chunks and 4.4M table chunks for OTT-QA and NQ, respectively)

\begin{table}[t!]
\centering
\begin{tabular}{lcc}
\toprule
\textbf{Dataset} & \textbf{OTT-QA} & \textbf{NQ} \\
\hline
Train & 41,469  & 79,168  \\
Dev & 2,214 & 8,757  \\
Test & 2,158 & 3,610  \\
\# Tables & 400K & 2.4M \\
\# Passages & 6.1M & 21M \\
\bottomrule
\end{tabular}
\caption{Statistics of Datasets}
\label{tab:stats}
\end{table}

\section{Training Details}
\label{sec:appendix-training}
We run all of our experiments once due to computation constraints, and we largely follow previous works' hyper-parameters settings. 
\subsection{Linker Training}
Only tables in OTT-QA train and dev set contain hyperlinks (8K for train, 0.8K for dev), and we only used these tables \footnote{\url{https://github.com/wenhuchen/OTT-QA}} to train our linker model (\S\ref{subsec:linker}).

To mine hard negatives for linker training, we adopted two strategies using BM25. For the first strategy, we used the entity mentions in the table are queries, and retrieve from an index of passage titles only. In the second one, we used the entity mentions along with the table title as queries, and retrieve from an index of passage titles concatenated with the first sentence from their corresponding pages. We observe very different negatives from these two strategies and we used negatives from both to train our linker model.  

During training, we use randomly sample one hard negative for every (mention, positive passage) pair, and also use in-batch negatives to compute the contrastive loss. We train entity mention proposal model for 40 epochs and table entity linking model for 100 epochs, both with batch size 64, learning rate 2e-5, and linear warm-up scheduler. 

\subsection{Retriever Training}
For training data, we directly used the data released by \citet{karpukhin-etal-2020-dense} for NQ \footnote{\url{https://github.com/facebookresearch/DPR}} and \citet{ma-etal-2022-open} for NQ-table-answerable set \footnote{\url{https://github.com/Mayer123/UDT-QA}}. Note that the NQ-table-answerable data is considered part of NQ in all settings. For OTT-QA, each question is annotated with one or more positive tables and we use BM25 to mine hard negatives from an index of all OTT-QA tables. 
Following previous work \cite{karpukhin-etal-2020-dense}, we train our retrievers (for both single-dataset and joint setting) for 40 epochs, with batch size 128, learning rate 2e-5 and select the best checkpoint using validation average rank on dev set. 

\subsection{Chainer Inference}
We set $\alpha=16$ and $\beta=9$ in \autoref{eqn:chainer} for OTT-QA, and $\alpha=10$ and $\beta=12$ for NQ. These hyper-parameters are selected based on the respective dev set's answer recall performance. We did a grid search over integer values in the range [1, 20] for both parameters. 

\subsection{Reader Training}
Following \citet{izacard-grave-2021-leveraging}, we train our reader models for 15000 steps, with batch size 64, learning rate 5e-5. 

\section{Results and Discussion}
\subsection{Impact of Linker \& Chainer}
\label{subsec:appendix-linker-chainer}
\begin{table}[t!]
\centering
\resizebox{\linewidth}{!}{
\begin{tabular}{lcccc}
\toprule
 & \textbf{R@20} &  \textbf{R@100} & \textbf{AR@20} &  \textbf{AR@50}\\
\midrule
OTT-QA Retriever & 82.7 & 92.5 & 31.2 & 37.5 \\
+Linker\&Chainer & 84.1 & 91.2 & 74.4 & 83.5 \\
\midrule
Retrieve full index & 82.3 & 91.8 & 32.7 & 40.2 \\
\bottomrule
\end{tabular}
}
\caption{Evidence Recall of OTT-QA single retriever on OTT-QA Dev set, where R@K evaluates gold table chunk recall and AR@K evaluates answer recall}
\label{tab:ott_retri_single}
\end{table}

\begin{table}[t!]
\centering
\resizebox{\linewidth}{!}{
\begin{tabular}{lcccc}
\toprule
 & \multicolumn{2}{c}{Dev} & \multicolumn{2}{c}{Test} \\
 & \textbf{AR@20} &  \textbf{AR@50} & \textbf{AR@20} &  \textbf{AR@50}\\
\midrule
Joint retriever & 82.3 & 87.0 & 83.6 & 88.2 \\
+Linker\&Chainer & 85.7 & 88.1 & 86.8 & 89.4 \\
\midrule
NQ Retriever & 83.2 & 87.4 & 84.1 & 88.4 \\
+Linker\&Chainer & 85.9 & 88.4 & 86.5 & 89.4 \\
\bottomrule
\end{tabular}
}
\caption{Answer Recall on NQ task}
\label{tab:nq_retri_single}
\end{table}

Here we evaluate the evidence recall performance of the OTT-QA single retriever in \autoref{tab:ott_retri_single}. We can see that compared to results in \autoref{tab:ott_retri}, the performance of the single dataset retriever is quite similar for all metrics, suggesting that a joint model is sufficient to learn multiple tasks under our framework. We conduct the same evaluation on NQ in \autoref{tab:nq_retri_single}. With the addition of Linker and Chainer, we see that the answer recall improved over the retriever-only setting. We also observe that the results for NQ-single retriever are quite similar to joint retriever, as seen on OTT-QA.

\section{Computation Time and Infrastructure}
We train our linker and retriever models on two A6000 GPUs, and the training took less than a day to finish. For the Chainer model, running inference for one million question-document pairs takes about 5 hours on one A100-80G GPU. We train our reader model on 16 V100-32G GPUS, which takes about 1 week to finish. 

Our linker model has ~330M parameters (3 bert-base encoders), the retriever model has ~220M parameters (2 bert-base encoders), the chainer model has 3B parameters and the reader model has 770M parameters.

\end{document}